\documentclass{article} 
\usepackage[preprint]{colm2026_conference}

\usepackage{microtype}
\usepackage{hyperref}
\usepackage{url}
\usepackage{booktabs}
\usepackage{enumerate}
\usepackage{amsthm}
\usepackage{graphicx}
\usepackage{epstopdf}
\usepackage{caption}
\usepackage{color}
\usepackage{enumitem}
\setlist[enumerate]{nosep} 
\usepackage{array}
\usepackage{tabularx}
\usepackage{multirow}
\usepackage{bm}
\usepackage{booktabs}
\usepackage{amssymb}
\usepackage{tabularx}
\newcolumntype{Y}{>{\centering\arraybackslash}X}

\usepackage{tcolorbox}
\tcbuselibrary{breakable}
\tcbuselibrary{skins}
\usepackage{caption}
\usepackage{subcaption}
\usepackage{listings}
\usepackage{wrapfig}
\usepackage{bbding}
\tcbuselibrary{skins}
\usepackage{colortbl}
\usepackage{fontawesome5}
\usepackage{authblk}
\usepackage{amsmath}
\usepackage{etoolbox}
\preto{\abstractkeywords}{\nolinenumbers} 

\usepackage{lineno}

\definecolor{darkblue}{rgb}{0, 0, 0.5}
\hypersetup{colorlinks=true, citecolor=darkblue, linkcolor=darkblue, urlcolor=darkblue}

\title{SWAA: Sliding Window Attention Adaptation for Efficient and Quality Preserving Long Context Processing}

\author[a]{Yijiong Yu}

\affil[a]{Oregon State University\\
    \texttt{\{yuyiji, huazheng.wang\}@oregonstate.edu}
}
\author[b,d]{Jiale Liu}
\author[b,d]{Qingyun Wu}
\affil[b]{Penn State University\\
    \texttt{\{jiale.liu, qingyun.wu\}@psu.edu}
}
\author[a,d]{Huazheng Wang}
\author[c]{Ji Pei}

\affil[c]{DeepSolution\\
    \texttt{research@deepsolution.chat}
}

\affil[d]{AG2ai, Inc.\\
    \texttt{{jiale, qingyun, huazheng}@ag2.ai}
}

\begin{document}
\maketitle

\begin{abstract}

The quadratic complexity of self attention in Transformer based LLMs renders long context inference prohibitively expensive. While Sliding Window Attention (SWA), the simplest sparse attention pattern, offers a linear complexity alternative, it suffers from catastrophic long context performance collapse, which stems from two fundamental factors: the training inference mismatch when naively applying SWA to models pretrained with Full Attention (FA), and the inherent structural inability to access distant information when applying SWA to every module at all times. To address these dual challenges, we propose Sliding Window Attention Adaptation (SWAA), a plug and play toolkit of recipes that adapts FA models to SWA without costly pretraining. SWAA systematically combines four core strategies to tackle these distinct issues: (1) Full Attention (FA) Decode and (2) Interleaving FA and SWA layers, which mitigate structural defects by selectively allowing access to distant information; alongside (3) preserving ``sink'' tokens and (4) lightweight fine tuning, which mitigate the training inference mismatch. Our experiments reveal that while isolated strategies are insufficient, specific synergistic combinations effectively recover long context performance. Despite varying computational overheads, our performance efficiency trade off analysis identifies optimal SWAA configurations for diverse scenarios, achieving 30\% to 100\% speedups for long context inference with acceptable quality retention. 
Our code, data and model weights are available at \href{https://github.com/yuyijiong/sliding-window-attention-adaptation}{github}.

\end{abstract}

\section{Introduction}

Transformer based Large Language Models (LLMs)~\citep{vaswani2017attention} demonstrate remarkable capabilities, but their self attention scales quadratically with the input sequence length, making long context processing inefficient. Sliding Window Attention (SWA), the most straightforward and widely adopted sparse attention pattern, restricts each token's attention to a fixed size local window. This reduces computational complexity to linearity, along with several other benefits (see Appendix \ref{app:drawback}). 

Despite its simplicity and ease of use, it causes catastrophic long context performance collapse due to two factors: the training inference mismatch, specifically the model's parameters cannot adapt to the drift of attention distribution, if directly applying SWA to models pretrained with Full Attention (FA); and the inherent structural difficulty of accessing the information of distant tokens~\citep{xiao2025sliding}, if SWA is applied to every module at all times, which structurally cuts off all direct information pathways to tokens beyond the local window.

Although various methods that use SWA for efficiency have attempted to address these dual challenges, current solutions struggle to balance flexibility and quality. For instance, training-free methods like streaming attention~\citep{xiao2024efficientstreaminglanguagemodels} offer high flexibility, enabling the rapid, plug and play application of SWA on arbitrary FA pretrained models by retaining ``sink tokens''. However, they struggle to guarantee long-context generation quality because the structural inaccessibility to distant information still exists. Conversely, models like Gemma2~\citep{gemmateam2024gemma2improvingopen} successfully mitigate this structural limitation by interleaving FA and SWA layers. Yet, this approach necessitates pretraining from scratch with this rigidly defined architecture, which is prohibitively costly, lacks flexibility, and restricts users' choices of base models. 

This dilemma provides our core motivation: \emph{Is it possible to leverage the efficiency of SWA to accelerate arbitrary LLMs, while simultaneously resolving these dual challenges at a low cost to maintain acceptable generation quality?}

We answer Yes to this question by proposing Sliding Window Attention Adaptation (SWAA), a plug and play toolkit of recipes for adapting FA pretrained models to SWA, which requires neither costly pretraining nor new modules beyond the standard Transformer architecture. Specifically, SWAA systematically combines four practical and composable strategies to tackle these distinct issues.

To mitigate SWA's structural inability to access distant information, we employ:
\begin{enumerate}
    \item \textbf{Full Attention (FA) Decode}: applying SWA selectively in time (only during the prefilling stage) while switching back to full attention for decoding, allowing the model to retrieve distant context when generating text. This strategy exhibits significant synergy with the use of Chain of Thought (CoT).
    \item \textbf{Interleaving FA/SWA layers}: applying SWA selectively in space by mixing full attention and SWA layers (e.g., applying SWA to half of the layers), ensuring distant information can still propagate through those FA layers.
\end{enumerate}

To mitigate the training inference mismatch caused by naively introducing SWA, we utilize:
\begin{enumerate}
    \setcounter{enumi}{2}
    \item \textbf{Keep First $k$ Tokens}: explicitly preserving attention to the first $k$ ``sink'' tokens, to keep the attention distribution stable.
    \item \textbf{Fine tuning with SWA}: lightweight, SWA aware supervised fine tuning on long context data.
\end{enumerate}

While FA Decode is a novel method we introduce, the other strategies have been proven individually effective in various contexts~\citep{xiao2024efficientstreaminglanguagemodels,gemmateam2024gemma2improvingopen,zhang2025lighttransferlongcontextllmsecretly}. However, how combine them for better, cheaper SWA adaptation remains unexplored.

In our experiments, we evaluate SWAA on Qwen3~\citep{yang2025qwen3technicalreport} and Llama3.1~\citep{grattafiori2024llama3herdmodels} across several long context benchmarks. First, we find that although each strategy is helpful, no single ingredient suffices to make SWA competitive with full attention in answer quality. Second, we show that specific synergistic combinations of these methods can effectively recover long context performance. Third, despite the varying degrees of computational overhead introduced by these methods, our analysis of performance-efficiency trade-offs identifies recommended SWAA configurations for diverse scenarios. For instance, SWAA can achieve nearly 100\% speedup with 90\% accuracy retention in efficiency first scenarios, or a 30\% speedup with nearly 100\% accuracy retention in quality first scenarios, making SWAA a flexible toolkit rather than a fixed, rigid strategy.

Our key contributions are:
\begin{itemize}
    \item We perform the first systematic study on adapting FA pretrained LLMs to SWA with multiple methods, specifically addressing both the training inference mismatch and the structural defects of SWA, providing a foundation for future research.
    \item We propose SWAA, a flexible toolkit of practical SWA adaptation recipes that offer a robust performance efficiency balance for various use cases, accelerating LLM inference from the bottom level.
    \item We implement our methods with Flash-Attention~\citep{dao2023flashattention2fasterattentionbetter} and vLLM~\citep{kwon2023efficientmemorymanagementlarge}, making SWAA plug and play and user friendly for practical deployment.
\end{itemize}

\section{Related Works}
\label{sec:related}
The $O(N^2)$ complexity of self attention in Transformers~\citep{vaswani2017attention} has spurred a wide field of research about more efficient language model architectures. Among the two most popular technological routes are sparse attention and linear attention.

\subsection{Sparse Attention}
Our work falls in this category. Sliding Window Attention (SWA) represents the most basic form of local sparse attention, yet its performance is inherently limited. Therefore, model architectures such as Longformer~\citep{beltagy2020longformerlongdocumenttransformer}, BigBird~\citep{zaheer2021bigbirdtransformerslonger}, and RATTENTION~\citep{wang2025rattentionminimalslidingwindow} combine local SWA on most tokens with special global attention on specific tokens to create a more powerful, albeit still sparse, pattern. Popular LLMs like Gemma2~\citep{gemmateam2024gemma2improvingopen} adopt SWA in half of their layers to balance the efficiency of SWA and performance of FA. Sliding Window Attention Training (SWAT)~\citep{fu2025slidingwindowattentiontraining} introduces architectural changes, such as sigmoid activation and balanced position embeddings, to stabilize SWA performance. More advanced methods like Native sparse attention~\citep{yuan-etal-2025-native}, although achieving excellent quality, involve more complicated implementation and optimization due to semantic aware attention operations (e.g., selecting the most important tokens based on attention weights).

Nevertheless, nearly all aforementioned methods require pretraining with a specific sparse pattern, which is costly and fails to leverage the advantages of existing pretrained models. Some works explore transforming a full attention trained models to sparse attention through post training or fine tuning. Deepseek sparse attention~\citep{deepseekai2025deepseekv32pushingfrontieropen} introduces a lightning indexer and a fine grained token selection mechanism to sparsely process tokens in long context, based on full attention trained Deepseek-V3.1, but still requires nearly 1T tokens for continued pre training. LightTransfer~\citep{zhang2025lighttransferlongcontextllmsecretly} attempts to adapt existing models to SWA without pretraining, which has the same motivation as ours. But it may generalize poorly across model families (see Appendix~\ref{app:lazy}).

\subsection{Linear Attention}

An alternative approach involves reformulating the attention mechanism entirely to achieve linear, $O(N)$, complexity. This includes methods such as RNN like linear attention transformers~\citep{katharopoulos_et_al_2020,peng2023rwkvreinventingrnnstransformer,sun2023retentivenetworksuccessortransformer} and structured state space models (SSMs) like Mamba~\citep{gu2024mambalineartimesequencemodeling}. Many works such as Jamba and Nemotron-Flash~\citep{lieber2024jambahybridtransformermambalanguage,noauthor_bamba_nodate,lingteam2025attentionmattersefficienthybrid, funemotron} interleave linear attention layers with traditional attention layers to create hybrid model structures. While promising, they represent a fundamental architectural departure from the standard Transformer, and thus must be trained from scratch. Meanwhile, their performance is generally weaker than traditional Transformer based LLMs, and may harm the math and reasoning ability.

\section{Method}
We incorporate four core adaptation methods into our SWAA framework to address the structural limitations and the training inference mismatch. Their specific operations and motivations are introduced below. 

\begin{figure}[ht]
  \centering
  \subfloat[FA Decode]{ \label{fig:presli}
    \includegraphics[height=0.25\columnwidth]{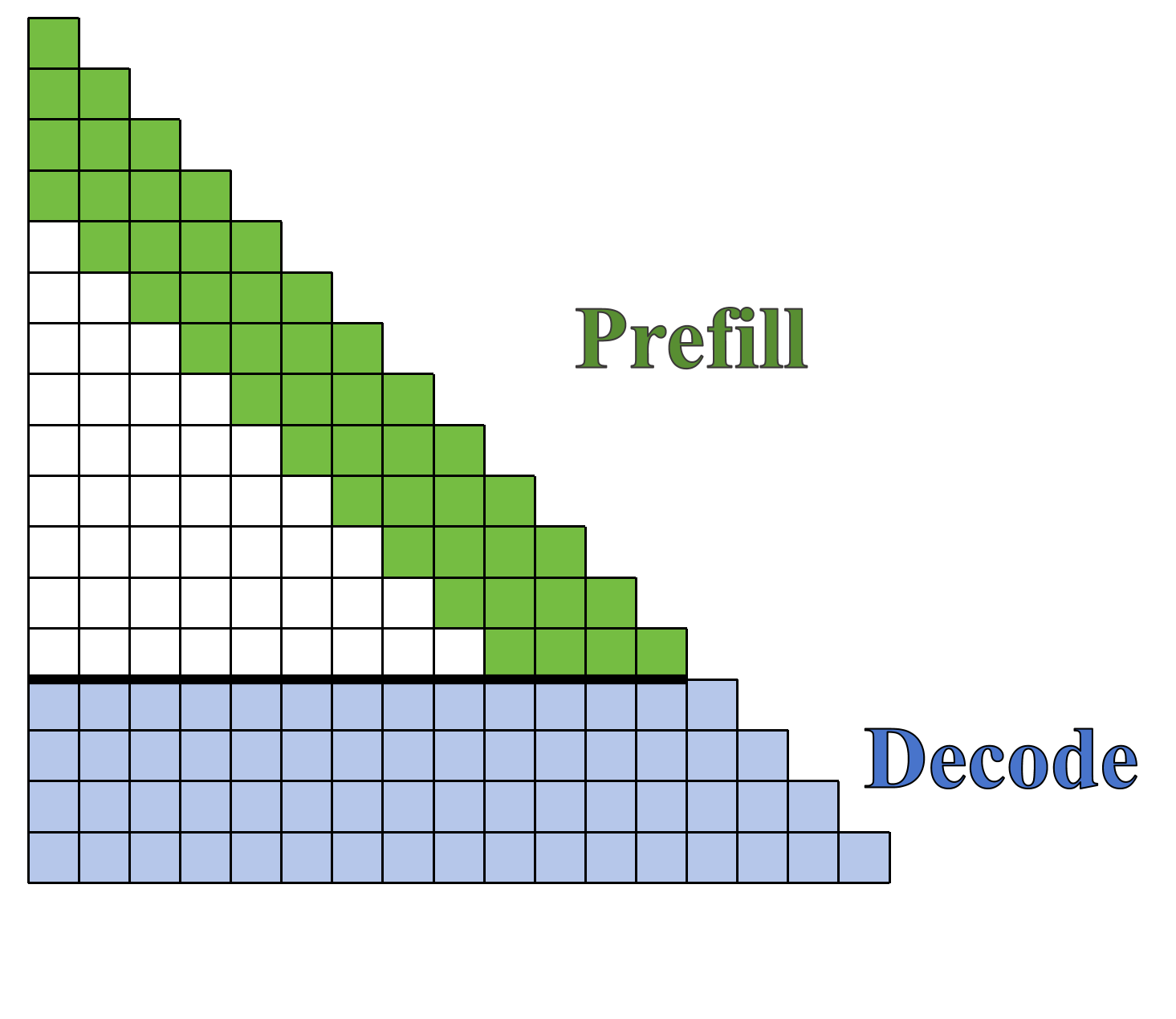}
    }
  \subfloat[Keep First]{ \label{fig:keepfirst} 
    \includegraphics[height=0.25\columnwidth]{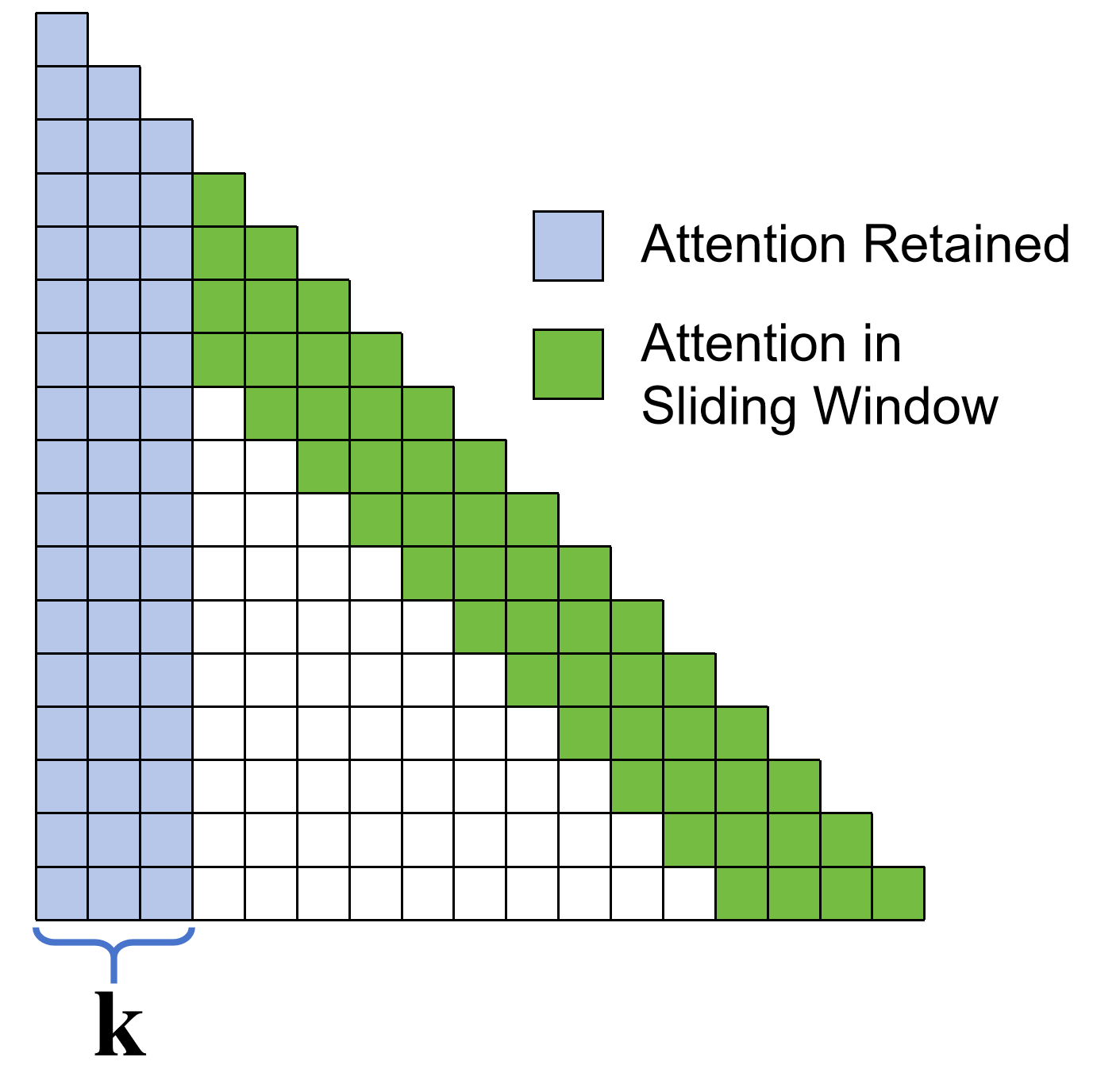}}
  \caption{(a) Attention mask for FA Decode. SWA is used for prompt tokens (prefill), and full attention is used for generated tokens (decode). (b) Attention mask for SWA combined with Keep First $k$ Tokens.}
\end{figure}

\subsection{Mitigating Structural Limitations}

\subsubsection{Full Attention (FA) Decode (Temporal Selection)}
\label{sec:fa}
This is a simple yet novel operation, which applies SWA \textbf{only} to the prefilling stage. During the decoding (auto regressive generation) stage, each token reverts to full attention, allowing the model to access all previous tokens across the entire long context. The resulting attention mask is depicted in Figure \ref{fig:presli}. 

This approach is inspired by human reading comprehension: humans typically scan a passage casually (prefilling) before thinking deeply to formulate an answer (decoding) for a specific problem. We term this strategy ``reading casually, thinking carefully.'' In our design, the SWA constrained prefilling stage corresponds to casual reading, while the full attention decoding stage enables careful thinking and unrestricted information retrieval.

Interestingly, this analogy implies that Chain of Thought (CoT) may naturally complements FA Decode, because enforcing an explicit ``thinking'' process naturally extends the generation length, thus significantly increases the chances and temporal pathways for the model to retrieve or reconstruct distant information that was structurally blocked during the prefilling stage.

\subsubsection{Interleaving FA/SWA Layers (Spatial Selection)}
\label{sec:interleave}
This method retains full attention on a subset of layers while applying SWA to the remainder, providing a spatial mechanism to propagate distant information through the network's depth while balancing the efficiency of pure SWA. A common strategy involves designating one in every $n$ layers to use full attention. For example, Gemma-2~\citep{gemmateam2024gemma2improvingopen} uses SWA only for layers [1, 3, 5, \dots], and Gemma-3~\citep{gemmateam2025gemma3technicalreport} uses SWA only for layers [5, 11, 17, \dots]. 

However, for an FA pretrained model, layers may inherently have distinct behaviors and functions, indicating this method may be sensitive to layer selection strategy. While methods like LightTransfer~\citep{zhang2025lighttransferlongcontextllmsecretly} have explored fine-grained layer selection, we find it is not consistently superior in practice when adapting existing FA models (see Appendix \ref{app:lazy}). Therefore, although the strategy better than fixed-interval selection exists in theory, there is currently no good method to ensure finding it. So, for simplicity and generalizability, we still adopt fixed-interval layer selections in our experiments, such as [0, 2, 4, \dots] or [1, 3, 5, \dots].

\subsection{Mitigating Training Inference Mismatch}

\subsubsection{Keep First $k$ Tokens}
Streaming attention~\citep{xiao2024efficientstreaminglanguagemodels} demonstrates that models pretrained with full attention allocate a disproportionate amount of attention to the initial tokens (the ``attention sink''). Naively applying SWA cuts off access to these sinks, causing severe mismatch. By permanently retaining attention to these initial $k$ tokens while using SWA, the stability of the attention distribution and the model's output is successfully maintained. As shown in Figure \ref{fig:keepfirst}, any subsequent token can attend to its local window \textbf{and} the initial $k$ tokens. 

Crucially, we extend prior approaches. While streaming attention operates only at the KV cache level (retaining the cache of sink tokens during decoding without accelerating prefilling), we directly customize the Flash-Attention-2~\citep{dao2023flashattention2fasterattentionbetter} kernel to implement this specific attention mask, so as to further accelerate the prefilling stage via SWA and eliminate the need for cumbersome KV cache modifications.

\subsubsection{Fine tuning}
Fine tuning is a highly direct approach to mitigating the training inference mismatch by updating the model's weights to adapt to sparse attention patterns. Crucially, fine tuning in our framework does not strictly imply training with naive SWA; rather, it can selectively incorporate any combination of the aforementioned methods. Integrating structural modifications, such as FA Decode, Interleaving Layers, or Keep First $k$ Tokens during the fine tuning phase is relatively straightforward.

However, enabling CoT (i.e., fine tuning a ``thinking'' model) presents a unique data challenge. Most available long context datasets provide only brief ground truth answers, which can be used to fine tune a standard instruct model, but is not applicable to a thinking model since it would severely degrade its reasoning style and capabilities. Since our goal is to \textit{restore} the model's original capabilities under SWA configurations rather than teach it new ones, we adopt a self distillation approach~\citep{yang2024selfdistillationbridgesdistributiongap} when fine tuning thinking models. Specifically, we utilize the original full attention thinking model to generate new answers for the dataset's questions, thereby perfectly preserving the CoT style if it is a thinking model. These generated answers are then filtered for correctness using GPT-5-Mini~\citep{noauthor_chatgpt_nodate} to construct our training dataset. For each question, we sample 4 answers with a temperature of $1.0$ to enhance diversity, which we empirically find yields slightly better adaptation than single answer generation.

\section{Experiment}
Our experiments are structured to first comprehensively analyze the accuracy of a wide variety of adaptation recipes, covering almost all possibly effective combinations of the components of SWAA, followed by a evaluation of their efficiency, rather than simply evaluating a single configuration where all proposed methods are activated simultaneously. This granular approach is driven by two primary considerations:

First, each structural method introduces varying degrees of computational overhead to LLM inference (as discussed in Appendix \ref{app:drawback}); indiscriminately adopting all of them may get suboptimal efficiency. Instead, we should evaluate multiple configurations to identify the optimal cost-effectiveness. 

Second, a systematic ablation by isolating and combining different elements allows us to rigorously verify the necessity and complex dynamics of every component, ultimately providing more comprehensive and useful insights for future research. 

\subsection{Experiment Setup}

\subsubsection{Models}

Our primary experiments use Qwen3-4B-Thinking and Qwen3-4B-Instruct~\citep{yang2025qwen3technicalreport}. The Thinking variant enforces chain of thought (CoT) style reasoning, whereas the Instruct variant usually just answers briefly, so we can clearly see the impact of CoT on overcoming SWA's structural defects. To ensure generality, we additionally evaluate Qwen3-30B-A3B-Thinking, Qwen3-30B-A3B-Instruct~\citep{yang2025qwen3technicalreport}, and Llama3.1-8B-Instruct~\citep{touvron2023llamaopenefficientfoundation}, as shown in Appendix \ref{app: other model}. All models are served with vLLM in \texttt{float16} precision using a batch size of 64, with our customized Flash-attention-2 kernel to support Keep First and FA Decoding. We use greedy decoding (temperature $=0$) for all evaluations. In preliminary experiments, we observed that vLLM yields slightly lower (about 1\% to 5\%) scores than HuggingFace Transformers due to precision related discrepancies.

\subsubsection{Evaluation Dataset}
SWA is identical to full attention when the context length is within the window size. Even if the model is fine tuned, we can simply disable the LoRA adapters for short prompts to get the same response as the original model. Therefore, our evaluations focus on long context benchmarks, as re-evaluating models on standard short context benchmarks (e.g., MMLU~\citep{hendrycks2021measuringmassivemultitasklanguage}, GPQA~\citep{rein2023gpqagraduatelevelgoogleproofqa}) is completely unnecessary. 

We select LongMemEval~\citep{wu2025longmemevalbenchmarkingchatassistants}, a benchmark consisting of various types of long context QA tasks with moderate difficulty, as our main benchmark, although it is originally designed for agent memory system evaluation. Its context length is controllable by selecting a specific number of chat sessions to concatenate as the context from a pool of hundreds of sessions (a session contains the chat history between user and assistant within a day). To create a moderately difficult and discriminative evaluation, we construct \textbf{LongMemEval\_24k} by sampling 10 sessions, resulting in 500 samples ranging from 16k to 32k with an average context length of 24k. 

We also experiment on LongBench-V2~\citep{bai2025longbenchv2deeperunderstanding} and Ruler~\citep{hsieh2024rulerwhatsrealcontext}, which are most commonly used long context benchmarks, for additional validation of generalizability. For LongBench V2, we retain only the samples whose context length is under 128k due to GPU memory limitations; thus, 311 of 500 samples are kept. For Ruler, we choose the Multi-Query task (the most challenging needle-in-a-haystack task), which contains 500 samples, and control the context length to 128k (counted by Qwen3 tokenizer). To judge the correctness for accuracy, we use LLM as judge with GPT-5-Mini~\citep{noauthor_chatgpt_nodate} for LongMemEval and exact match for LongBench-V2 and Ruler. However, we find other LongBench-V2 is too difficult for 4B level models, and Ruler's task types are not diverse enough (see details in Appendix \ref{app:bench}), so our analysis will more focus on the clearer results of LongMemEval.

\subsubsection{Training Details}
For the fine tuning dataset, we initially considered LongAlign~\citep{bai2024longalignrecipelongcontext}, a widely used long context fine tuning dataset for adapt a regular length model to long context tasks. However, since its sample count ($\sim$10,000) is insufficient, we incorporate an additional 6,000 samples from Fusang-v1-long~\citep{Fusang-V1}, a more comprehensive corpus of over 40,000 long context samples that includes LongAlign as a subset. We perform SWA aware fine tuning using LoRA~\citep{hu2021loralowrankadaptationlarge}. Unless otherwise noted, we use rank $r=16$ and $\alpha=128$, and apply LoRA only to the query, key, and value projection modules. We adopt this parameter efficient setting because full parameter fine tuning often leads to overfitting and degradation of the model's original capabilities in our preliminary experiments. We use a learning rate of 1e-4 with a cosine decay schedule. Models are fine tuned for a single epoch on the sampled long context dataset since we observe no meaningful gains from additional epochs (see Appendix~\ref{app:epoch}). Training of each SWAA configuration takes approximately 12 hours on an 8*H20 GPU server for Qwen3-4B and 30 hours for Qwen3-30B-A3B.

\begin{table*}[t]
\centering
\caption{The accuracy of Qwen3-4B on LongMemEval, LongBench V2, and Ruler with different SWAA recipes.}
\label{tab:merged_4b}

\setlength{\tabcolsep}{2pt}
\begin{tabular}{l|cclcc|cc|cc|cc}
\toprule
\multirow{2}{*}{\textbf{No.}} & \multirow{2}{*}{\textbf{SFT}} & \textbf{Window} & \multirow{2}{*}{\textbf{FA layers}} & \textbf{Keep} & \textbf{FA} & \multicolumn{2}{c|}{\textbf{LongMem}} & \multicolumn{2}{c|}{\textbf{LB V2}} & \multicolumn{2}{c}{\textbf{Ruler}} \\
 & & \textbf{size} & & \textbf{first} & \textbf{decode} & \textbf{Think} & \textbf{Non T} & \textbf{Think} & \textbf{Non T} & \textbf{Think} & \textbf{Non T} \\
\midrule
0 & False & Full & / & / & / & \textbf{73.0} & \textbf{62.0} & \textbf{34.6} & \textbf{35.2} & \textbf{85.6} & \textbf{92.8} \\
1 & False & 2k & [] & 0 & False & 3.2 & 11.0 & 9.4 & 25.8 & 0.0 & 0.0 \\
2 & False & 8k & [] & 0 & False & 13.2 & 19.8 & 15.1 & 22.1 & 0.0 & 0.0 \\
\midrule
3 & False & 2k & [] & 10 & False & 16.0 & 15.6 & 7.7 & 25.8 & 0.0 & 0.0 \\
4 & False & 2k & [] & 0 & True & 11.8 & 14.2 & \textbf{26.2} & 25.2 & 0.2 & 0.0 \\
5 & False & 2k & [1, 3, 5, ...] & 0 & False & 13.4 & 18.4 & 12.1 & 23.5 & 0.0 & 0.0 \\
6 & False & 8k & [] & 0 & True & 26.2 & 25.0 & 22.8 & 25.5 & 0.0 & 0.0 \\
\midrule
7 & False & 2k & [] & 10 & True & 38.2 & 20.6 & 25.8 & 25.2 & 1.0 & 0.0 \\
8 & False & 2k & [] & 100 & True & 50.0 & 17.8 & 24.2 & 26.5 & 3.2 & 0.4 \\
9 & False & 2k & [] & 1000 & True & 50.0 & 20.2 & 23.8 & 25.2 & 3.2 & 0.2 \\
10 & False & 2k & [0, 2, 4, ...] & 10 & False & 17.0 & 14.8 & 19.8 & 29.2 & 0.0 & 0.0 \\
11 & False & 2k & [0, 2, 4, ...] & 0 & True & 32.2 & 26.0 & 23.8 & 29.9 & 1.0 & 2.0 \\
12 & False & 2k & [1, 3, 5, ...] & 10 & False & 25.8 & 36.4 & 21.1 & 29.9 & 0.0 & 0.0 \\
13 & False & 2k & [1, 3, 5, ...] & 0 & True & 59.2 & 34.8 & 28.5 & 26.8 & 22.8 & 32.8 \\
14 & False & 4k & [] & 10 & True & 38.0 & 24.4 & 27.9 & 27.5 & 1.6 & 2.2 \\
15 & False & 8k & [] & 10 & True & 49.2 & 35.2 & \textbf{36.7} & 30.2 & 3.2 & 6.2 \\
\midrule
16 & False & 2k & [0, 2, 4, ...] & 10 & True & 36.0 & 17.2 & 30.5 & 30.2 & 10.0 & 3.0 \\
17 & False & 2k & [1, 3, 5, ...] & 10 & True & 65.0 & \textbf{53.6} & \textbf{35.4} & 28.6 & \textbf{59.2} & 75.8 \\
18 & False & 2k & [1, 3, 5, ...] & 100 & True & \textbf{68.8} & 50.6 & \textbf{35.6} & \textbf{31.8} & 58.0 & \textbf{78.8} \\
19 & False & 2k & [1, 5, 9, ...] & 10 & True & 53.2 & 31.4 & 29.6 & 26.4 & 9.8 & 26.4 \\
20 & False & 2k & [1, 9, 17, ...] & 10 & True & 36.4 & 18.8 & 28.6 & 25.4 & 0.6 & 0.2 \\
21 & False & 2k & [3, 7, 11, ...] & 10 & True & 54.2 & 34.6 & 27.7 & 26.0 & 1.6 & 4.2 \\
22 & False & 4k & [1, 3, 5, ...] & 100 & True & \textbf{73.0} & \textbf{54.2} & 33.4 & \textbf{31.2} & \textbf{65.4} & \textbf{83.6} \\
23 & False & 8k & [1, 3, 5, ...] & 100 & True & \textbf{71.6} & \textbf{56.6} & \textbf{35.4} & \textbf{31.5} & \textbf{67.2} & \textbf{86.0} \\
\midrule
\midrule
24 & True & Full & / & / & / & \textbf{74.6} & \textbf{63.4} & \textbf{37.9} & \textbf{34.9} & \textbf{88.2} & \textbf{90.6} \\
\midrule
25 & True & 2k & [] & 0 & False & 18.8 & 23.8 & 7.4 & 30.9 & 0.0 & 0.0 \\
\midrule
26 & True & 2k & [] & 100 & False & 15.6 & 24.0 & 6.0 & 30.2 & 0.0 & 0.0 \\
27 & True & 2k & [] & 0 & True & 57.9 & 42.0 & 29.2 & 30.2 & 0.4 & 0.8 \\
28 & True & 2k & [1, 3, 5, ...] & 0 & False & 63.6 & \textbf{54.6} & 29.5 & 31.9 & \textbf{59.0} & \textbf{62.2} \\
\midrule
29 & True & 2k & [] & 10 & True & 59.0 & 41.4 & 28.9 & 29.2 & 9.2 & 14.4 \\
30 & True & 2k & [] & 100 & True & 62.2 & 42.6 & 29.2 & 30.5 & 13.4 & 20.4 \\
31 & True & 2k & [1, 3, 5, ...] & 0 & True & \textbf{73.2} & \textbf{58.8} & \textbf{38.3} & \textbf{34.6} & \textbf{74.0} & \textbf{85.8} \\
32 & True & 2k & [0, 2, 4, ...] & 0 & True & 66.0 & $\backslash$ & 31.5 & $\backslash$ & 8.0 & $\backslash$ \\
33 & True & 2k & [1, 5, 9, ...] & 0 & True & \textbf{68.8} & 47.0 & 32.0 & \textbf{32.2} & 7.6 & 28.0 \\
\midrule
34 & True & 2k & [1, 3, 5, ...] & 100 & True & \textbf{73.2} & \textbf{61.4} & \textbf{37.2} & \textbf{33.9} & \textbf{72.8} & \textbf{89.2} \\
\bottomrule
\end{tabular}
\end{table*}
\subsection{Experiment Results}

\subsubsection{Accuracy Analysis}

A macroscopic analysis of our experimental results of accuracy (Table~\ref{tab:merged_4b}) reveals a fundamental principle: the catastrophic degradation caused by SWA cannot be resolved by addressing the dual bottlenecks, training inference distribution drift and structural severing of distant semantic flow, in isolation. Applying any single method, such as Keep First, FA Decode, Interleaving Layers, or naive SFT, yields negligible accuracy improvements over pure SWA (Table~\ref{tab:merged_4b}, rows 3 to 5 and 25). Conversely, when we pair methods targeting distinct challenges (e.g., Fine tuning and FA Decode, Fine tuning and Interleaving layers, or Keep First and FA Decode), we observe significant accuracy recoveries.

\textbf{FA Decode and Temporal Retrieval Synergy.} Applying SWA strictly during prefilling while reverting to FA during decoding provides a temporal selection mechanism that temporarily reopens direct pathways to distant tokens. We find this structural mitigation exhibits a profound synergy with the use of CoT, i.e. thinking. Because CoT naturally extends the auto regressive generation phase, it provides the model with more sequential steps to actively retrieve and integrate the broad context that was structurally under processed during the prefilling stage. This temporal retrieval synergy is validated by the superior performance of the thinking model over the standard instruct variant under FA Decode configurations (rows 13 and 18).

\textbf{Interleaving FA and SWA Layers: Spatial Information Highways.} While FA Decode addresses temporal bottlenecks, interleaving FA layers establishes vital spatial pathways, allowing distant information to propagate through the network's depth. This is a highly potent structural mitigation, but zero shot adaptation is sensitive to layer selection. For instance, configuring odd numbered layers with FA vastly outperforms even numbered ones for Qwen3-4B (row 13 vs. row 11). This asymmetry suggests that specific layers in pretrained LLMs inherently function as global routing hubs, and misaligning SWA with these hubs disrupts the semantic flow.

\textbf{Synergy of Structural Methods.} Interestingly, combining two purely structural interventions, FA Decode and Interleaving Layers, also delivers substantial gains (row 13). This occurs because structural improvements selectively enable SWA, naturally nudging the overall attention pattern closer to its original full attention distribution. Consequently, this implicitly alleviates the training and inference mismatch alongside resolving structural deficits.

\textbf{Keep First $k$ Tokens: Training free Attention stabilizer.} Models pretrained with Full Attention (FA) develop a strong reliance on initial ``sink'' tokens. Prior to fine tuning, naively applying SWA cuts off access to these sinks, causing severe attention distribution drift. Preserving attention to these first $k$ tokens acts as a crucial distributional anchor, preventing complete generation collapse (rows 7 and 9). However, its effect is obviously weaker than, and will be covered by SFT, which intrinsically reshape the attention weights to not relying heavily on these sinks, rendering the Keep First strategy largely optional (row 27 vs. 29).

\textbf{Fine Tuning: Aligning Model Weights with Sparse Attention Patterns.} While training free methods provide partial recovery, SFT acts as the ultimate unifier, parametrically aligning the model's internal representations with the modified sparse attention structures. Combining structural methods (FA Decode or Interleaving Layers) with lightweight fine tuning virtually eliminates the performance gap, restoring accuracy to perfectly match the full attention baseline (rows 27, 28, and 31).

The above findings are mainly based on the results of LongMemEval and LongBench V2, which are more general. While for Ruler, as it involves only the needle in a haystack task, it exhibits a more extreme pattern: the accuracy can hardly be improved unless we enable Interleaving Layers, coupled with at least one another method (row 13 and 28). This indicates the retrieval behavior may be concentrated in certain specific layers, therefore these layers must retain full attention.

\begin{figure*}[htb]
  \centering
  \subfloat[Qwen3-4B-Thinking]{
    \includegraphics[height=0.46\columnwidth]{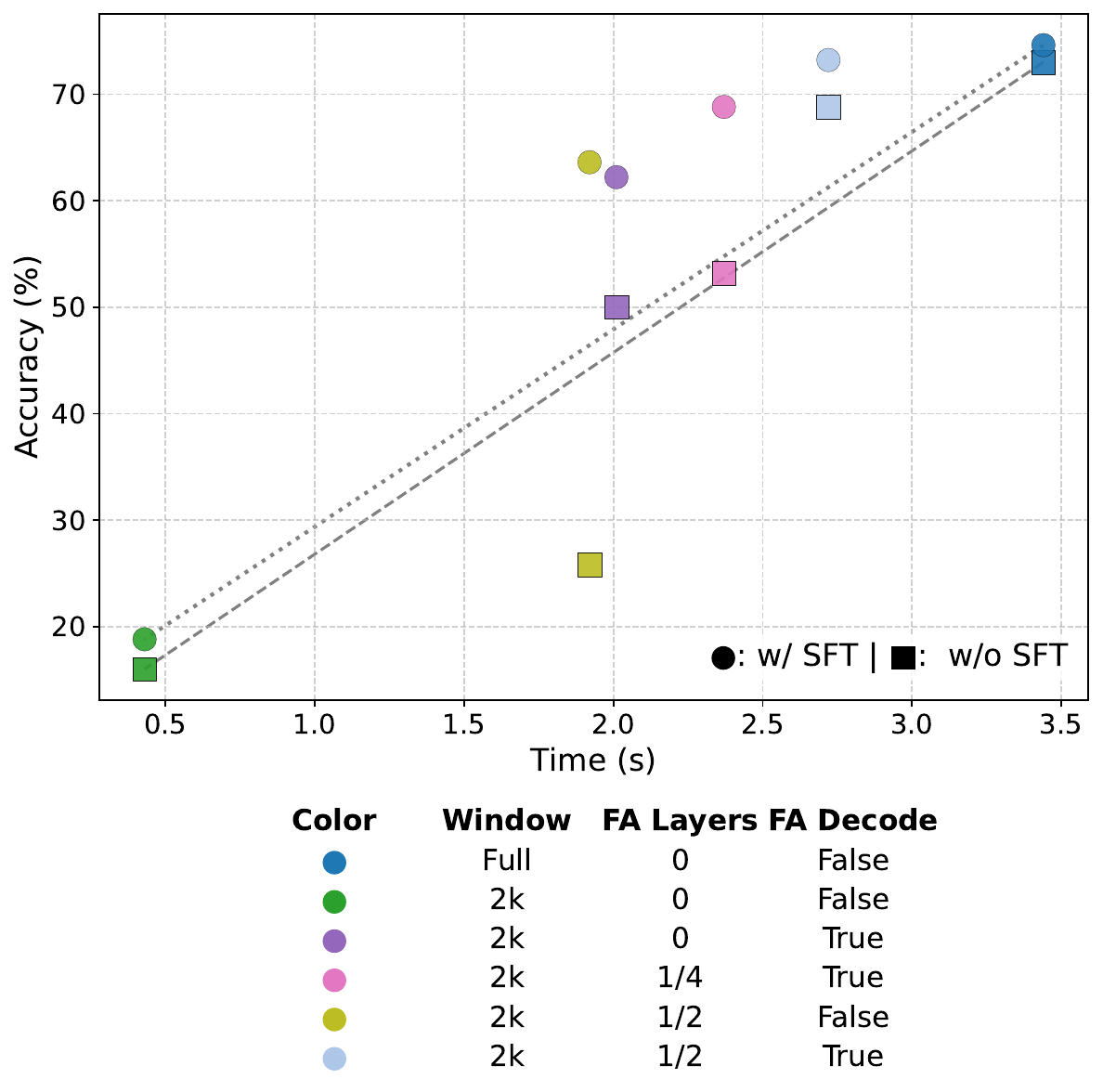}}
    \hspace{0.1em}
  \subfloat[Qwen3-4B-Instruct]{
    \includegraphics[height=0.46\columnwidth]{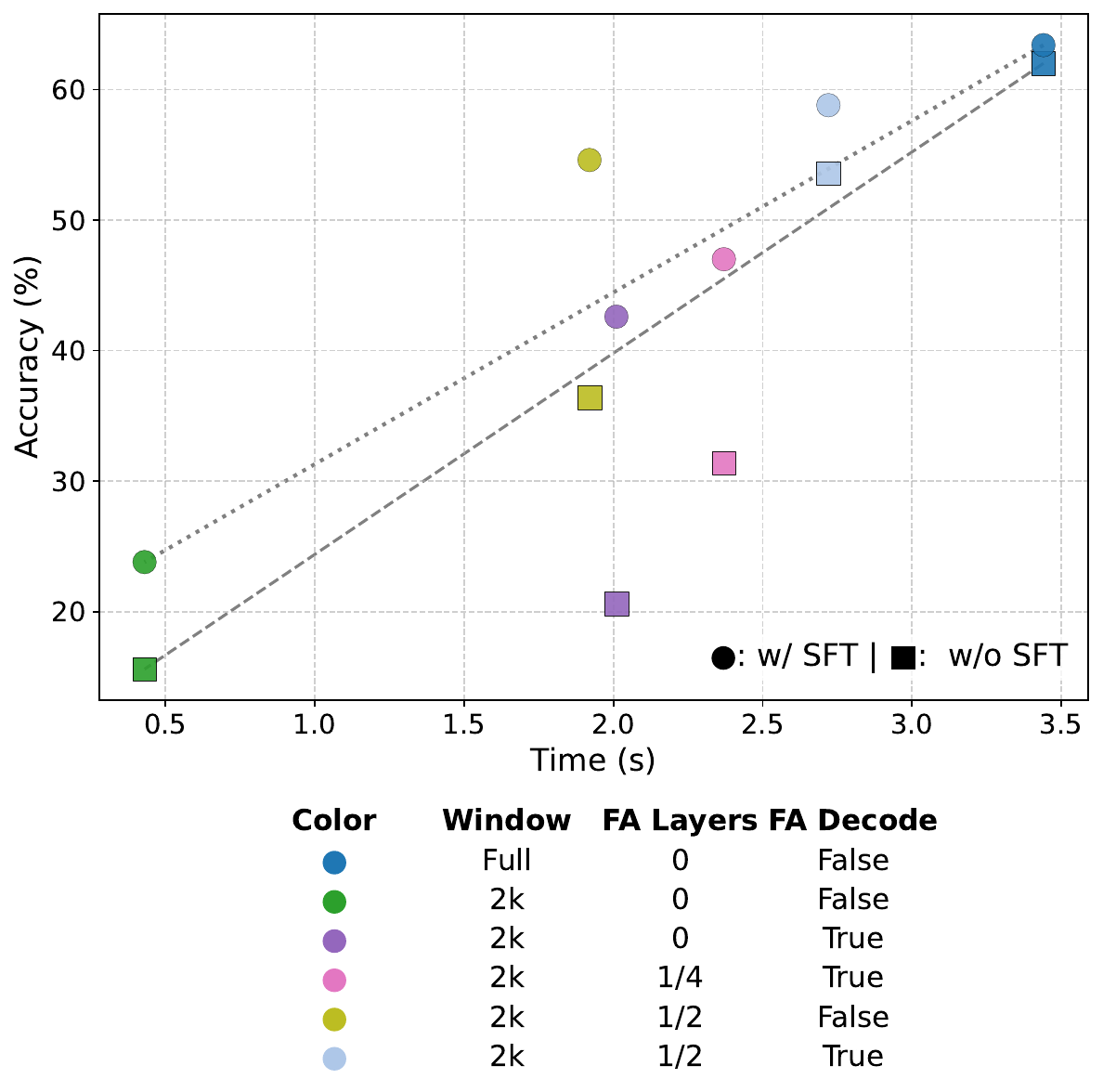}}
  \caption{Accuracy and inference time of each configuration of Qwen3-4B on LongMemEval}
    \label{fig:time}
\end{figure*}

\subsubsection{Efficiency Analysis and Recommended Recipes}

Although every method is helpful to accuracy recovery, Interleaving Layers and FA Decode introduce varying degrees of computation overhead in inference compared to pure SWA, indicating that the performance-efficiency trade-off of each SWAA configurations must also be evaluated, to find the most cost-effective one. So, we evaluate the average running time per request of Qwen3-4B-Thinking on a single H100 GPU using vLLM's \texttt{bench\_serve} utility~\citep{kwon2023efficientmemorymanagementlarge} with random input data and a total of 100 requests. The prompt length and output length are set to 128k and 512 tokens, respectively, representing a typical long context QA setting. More specific metrics such as time to first token (TTFT), time per output token (TPOT) and total throughput are shown in Appendix~\ref{app:speed}.

To visualize the performance efficiency trade off, Figure~\ref{fig:time} plots each configuration’s accuracy on LongMemEval\_24k~\citep{wu2025longmemevalbenchmarkingchatassistants} against its average running time, while detailed TTFT, TPOT, and throughput statistics for each configuration are provided in Appendix~\ref{app:speed}. We draw a line between the full attention point and the naive SWA point as a baseline curve: configurations above this line offer a better accuracy latency balance intuitively. For configurations with nearly identical time costs, we display only the one with the highest accuracy. Since \textbf{Keep First $k$} has negligible impact on runtime (Appendix~\ref{app:speed}), all plotted configurations fix $k=10$.

\begin{table}[t]
\centering
\caption{Recommended SWA adaptation recipes for different needs and scenarios: whether you want to train, use a thinking model, or prefer efficiency or quality. \faIcon[regular]{minus-square} means optional.}
\setlength{\tabcolsep}{1pt}
\small
\begin{tabularx}{\linewidth}{YYY|YYY}
\toprule
 \textbf{Train} & \textbf{Thinking} & \textbf{Priority} &  \textbf{FA Decode} & \textbf{Inter-leaving} & \textbf{Keep First} \\
\midrule
 No & No & Any & \faIcon[regular]{check-square} & \faIcon[regular]{check-square} & \faIcon[regular]{check-square}  \\
 No & Yes & Efficiency & \faIcon[regular]{check-square} &  \faIcon[regular]{window-close} & \faIcon[regular]{check-square}  \\
  No & Yes & Quality & \faIcon[regular]{check-square} &  \faIcon[regular]{check-square}  & \faIcon[regular]{check-square}  \\
  
\midrule
  Yes & Any & Efficiency & \faIcon[regular]{window-close} &  \faIcon[regular]{check-square}  & \faIcon[regular]{minus-square}  \\ 
  Yes & Any & Quality & \faIcon[regular]{check-square} &  \faIcon[regular]{check-square}  & \faIcon[regular]{minus-square}  \\
  
  \bottomrule
\end{tabularx}

\label{tab:rec}
\end{table}

From Figure \ref{fig:time}, we observe that many configurations achieve a clearly better performance efficiency ratio than baselines. For example, after SFT, (1) enabling either FA decoding or Interleaving Layers can enhance the inference speed by almost 100\%, while maintaining nearly 90\% of the original accuracy; (2) while combining them can achieve nearly 100\% original accuracy maintenance and over 30\% acceleration; (3) By controlling the number of layers using FA, efficiency and quality can be achieved between that of (1) and (2). Furthermore, for the thinking model, more points lie above the baseline curve compared to the non thinking model, indicating that \textbf{CoT} generally has a positive effect on improving the performance efficiency ratio of SWAA.

Thus, we conclude that many SWAA configurations reach excellent performance efficiency trade offs, but there is no single metric to quantify such trade offs to decide the globally optimal one. We therefore summarize \textbf{recommended SWA adaptation recipes} tailored to various deployment scenarios in Table~\ref{tab:rec}. We must note that specific parameters should be flexibly set to meet application specific requirements, without the need to follow our experimental parameters (e.g., a 2k window, $k=10$). For example, users can increase the window size to 4k or $k$ to 128 for higher accuracy and acceptable additional overhead.

\section{Conclusion}

In this work, we validate the feasibility of adapting full attention pretrained LLMs to Sliding Window Attention (SWA) for better efficiency, offering a cost effective alternative, SWAA, that avoids training sparse attention models from scratch. By systematically deconstructing the adaptation process, we identify that the catastrophic degradation observed in naive implementations can be effectively mitigated through synergistic combinations of auxiliary methods. Our extensive experiments across Qwen and Llama families demonstrate that while trade offs between computational overhead and model performance are inevitable, optimized configurations can achieve excellent performance efficiency balance to effectively accelerate LLM inference.




\bibliography{colm2026_conference}
\bibliographystyle{colm2026_conference}
\appendix
\section{SWA's Benefits and Each Method's Drawbacks}
\label{app:drawback}

SWA reduces the computational complexity to $O(N \cdot W)$, where $W$ is the window size. The benefits are threefold: (1) SWA reduces the computational load, (2) conserves GPU memory by limiting the required Key-Value (KV) cache, and (3) enhances KV cache reusability beyond traditional prefix caching, since a token's state is independent of tokens outside its local window. However, there is no free lunch: to preserve long-context capability, we have to use the methods of SWAA, each of which has drawbacks, impairing the aforementioned benefits to varying degrees.

\textbf{FA Decode} presents two primary drawbacks: (1) the benefits apply only to prefilling, while decoding speed is not accelerated as it utilizes full attention, and (2) the GPU memory required for the KV cache is not reduced, as the KV cache for the full context must be retained for decoding. Using CoT further increases the decoding time, reducing the speedup ratio. In practice, however, many distributed LLM services have to recompute the KV cache of the entire chat history because storing and loading the KV cache complicates engineering systems, making prefilling occurs more frequently than expected, thereby amplifying the advantage of this method.

\textbf{Interleaving Layers} introduces the most significant overhead, as only a subset of layers benefits from the computational savings of SWA. Furthermore, the GPU memory required for the KV cache is not reduced for the full-attention layers. Additionally, this method negates the KV cache reusability advantage of SWA, as the existence of full-attention layers violates the independence of the KV cache beyond the local window.

\textbf{Keep First} introduces very minor computational overhead, but it complicates efficient KV cache reuse. Due to positional encoding, a token's KV state depends on its position relative to the initial $k$ tokens, hindering simple cache reuse across different requests. A position encoding separation or offsetting mechanism may be needed.


\section{Some Problems of Other Long-context Benchmarks}
\label{app:bench}
We find existing long-context benchmarks problematic for our specific needs. For example: 
\begin{enumerate}
\item LongBench~\citep{bai2024longbenchbilingualmultitaskbenchmark} is classic and widely used, but its average context length (most are under 16k) is relatively short for modern models, i.e., it is already too easy. And its data source is too old, leading to a risk of test data leakage. So we choose not to use it.

\item Ruler~\citep{hsieh2024rulerwhatsrealcontext} has controllable context length, but its tasks are almost all synthetic and most of them are needle-retrieval tasks, thus failing to reflect the model's overall long-context capability in real-world scenarios. 

\item LongBench-V2~\citep{bai2025longbenchv2deeperunderstanding} is well-designed to necessitate deep understanding and reasoning over very long context. But it is too challenging for 4B-level models (e.g., Qwen3-4B-Thinking only gets 35\% accuracy, which is too close to the random guessing baseline of 25\%), making the improvement of different methods less distinguishable. Moreover, since it is in a multiple-choice question format, the results may not be sufficiently reliable because the model has a 25\% chance of guessing the correct option.

However, despite these drawbacks, they have been widely used in long-context model benchmarking. Thus we still elect to conduct our experiments on LongBench V2~\citep{bai2025longbenchv2deeperunderstanding} and Ruler~\citep{hsieh2024rulerwhatsrealcontext} to verify the generalizability of our conclusions, as shown in Appendix \ref{app: other model}.

\end{enumerate}

\section{Experiment Results of Other Models}
\label{app: other model}

We show the accuracy results of Qwen3-30B-A3B in Table \ref{tab:merged_30b}, and the results of Llama3.1-8B-Instruct in Table \ref{tab:merged_llama}. Due to the time-intensive nature of training, we only test a selective set of configurations with fine-tuning. For Llama3.1-8B-Instruct on Ruler, the context length is controlled to 32k due to its weaker long-context capability. From the results, we observe that for any benchmark, the trend of accuracy change is mostly consistent with our previous conclusions, demonstrating their generalizability.

\section{Inference Efficiency}
\label{app:speed}

The TTFT, TPOT and total throughput when using vLLM on a single H100 are shown in Table \ref{tab:speed}. Since inference speed is highly dependent on hardware, implementation details, and workload characteristics, these numbers should be interpreted as reference values. From the results, we can still conclude that:
\begin{enumerate}
\item Interleaving Layers and FA Decode significantly slow down the speed compared to pure SWA.

\item Keep First $k$ Tokens has a negligible impact on efficiency. 

\item Increasing the window size slightly increases inference time. For example, increasing from 2k to 4k decreases throughput by only 10\%, but a 4k window generally achieves higher accuracy based on previous experiments. Therefore, in practice, a 4k window is a more common choice.

\end{enumerate}

In theory, FA Decode should yield a decoding speed identical to that of full attention. Yet, in this table, we observe acceleration on TPOT. This is because vLLM-v1 typically mixes different requests' prefilling and decoding tokens in one sequence to improve GPU utilization. Thus, the speeds of prefilling and decoding may affect each other. If processing only a single request, the situation differs. For example, when the generation length is set to 2000, we find decoding takes over 95\% of the total time, rendering the acceleration of the prefilling stage negligible—i.e., SWA with FA Decode is almost unable to improve efficiency in such extreme cases.

\section{Influence of Training Epochs}
\label{app:epoch}
As shown in Table \ref{tab:epoch}, training for more than 1 epoch yields no improvement. Therefore, we choose to train for only 1 epoch.

\section{Results of LightTransfer}
\label{app:lazy}

LightTransfer~\citep{zhang2025lighttransferlongcontextllmsecretly} represents a promising attempt at SWA adaptation on full-attention models without pretraining. It proposes a layer selection method for SWA adaptation that calculates a "lazy ratio," represented by the ratio of attention from tokens at the end of the sequence (from a calibration dataset) to recent tokens versus global tokens. Layers with a higher "lazy ratio" are selected to apply SWA, while the rest retain full attention. This method is intuitive and theoretically sound, but our experiments reveal some negative results.


Since the complete code of LightTransfer is not open-source, we reproduce this method using LongAlign~\citep{bai2024longalignrecipelongcontext} as the calibration dataset for lazy layer detection, where the number of last tokens is set to 64, and the recent token window is set to 1024. From our experimental results shown in Table \ref{tab:lazy}, we find that:
\begin{enumerate}
\item For Qwen3-4B, LightTransfer even yields a counterproductive effect; allowing lazy layers to use FA yields higher scores, while following the original method (letting non-lazy layers use FA) results in significantly lower scores.
\item For Qwen3-30B, it provides nearly no improvement over fixed-interval selection.
\item Only for Llama3.1-8B does LightTransfer show advantages.
\end{enumerate}

Therefore, we conclude that LightTransfer does not yield stable performance across various models. Although fine-grained layer selection methods are theoretically superior, we believe they require further investigation before integration into our SWAA recipes.

\begin{table*}[ht]
\centering
\caption{The accuracy of Qwen3-30B-A3B-Thinking and Qwen3-30B-A3B-Instruct on LongMemEval, LongBench-V2, and Ruler}
\label{tab:merged_30b}

\setlength{\tabcolsep}{2pt}
\begin{tabular}{l|cclcc|cc|cc|cc}
\toprule
\multirow{2}{*}{\textbf{No.}} & \multirow{2}{*}{\textbf{SFT}} & \textbf{Window} & \multirow{2}{*}{\textbf{FA layers}} & \textbf{Keep} & \textbf{FA} & \multicolumn{2}{c|}{\textbf{LongMem}} & \multicolumn{2}{c|}{\textbf{LB-V2}} & \multicolumn{2}{c}{\textbf{Ruler}} \\
 & & \textbf{size} & & \textbf{first} & \textbf{decode} & \textbf{Think} & \textbf{Non-T} & \textbf{Think} & \textbf{Non-T} & \textbf{Think} & \textbf{Non-T} \\
\midrule
0 & False & Full & / & / & / & \textbf{79.2} & \textbf{71.6} & \textbf{49.7} & \textbf{42.6} & \textbf{95.6} & \textbf{99.4} \\
1 & False & 2k & [] & 0 & False & 0.0 & 0.4 & 0.0 & 0.0 & 0.0 & 0.0 \\
2 & False & 8k & [] & 0 & False & 0.0 & 0.2 & 0.0 & 0.0 & 0.0 & 0.0 \\
\midrule
3 & False & 2k & [] & 10 & False & 0.0 & 2.8 & 9.1 & 32.2 & 0.0 & 0.0 \\
4 & False & 2k & [] & 0 & True & 0.2 & 0.2 & 0.0 & 0.0 & 0.0 & 0.0 \\
5 & False & 2k & [0, 2, 4, ...] & 0 & False & 21.0 & 28.4 & 20.1 & 25.8 & 1.4 & 17.0 \\
\midrule
6 & False & 2k & [] & 10 & True & 43.8 & 23.6 & 9.1 & 32.2 & 0.0 & 0.0 \\
7 & False & 2k & [] & 100 & True & 58.6 & 22.2 & 10.4 & 28.2 & 2.4 & 3.0 \\
8 & False & 2k & [] & 1k & True & 59.0 & 25.4 & 11.7 & 29.5 & 3.4 & 5.8 \\
9 & False & 4k & [] & 10 & True & 49.8 & 26.6 & 26.8 & 30.9 & 3.0 & 1.2 \\
\midrule
10 & False & 2k & [0, 2, 4, ...] & 10 & True & \textbf{74.8} & \textbf{63.0} & 22.1 & \textbf{33.6} & 85.4 & 98.8 \\
11 & False & 2k & [0, 4, 8, ...] & 10 & True & 48.8 & 23.8 & 12.4 & 29.5 & 1.6 & 1.8 \\
12 & False & 2k & [1, 3, 5, ...] & 10 & True & 51.6 & 24.0 & \textbf{30.2} & 28.9 & 17.8 & 64.2 \\
13 & False & 2k & [2, 6, 10, ...] & 10 & True & 64.8 & 44.2 & 21.1 & 35.6 & 35.8 & 50.2 \\
14 & False & 4k & [0, 2, 4, ...] & 100 & True & \textbf{74.6} & \textbf{64.4} & \textbf{29.5} & \textbf{35.9} & 89.4 & 99.0 \\
\midrule
\midrule
15 & True & Full & / & / & / & \textbf{79.6} & \textbf{72.0} & \textbf{43.6} & \textbf{42.0} & \textbf{97.6} & \textbf{99.4} \\
\midrule
16 & True & 2k & [] & 0 & True & 62.2 & 51.0 & 35.9 & 33.9 & 4.6 & 22.4 \\
\midrule
17 & True & 2k & [] & 100 & True & 65.6 & 50.8 & \textbf{36.6} & 32.9 & 10.8 & 57.8 \\
18 & True & 2k & [0, 2, 4, ...] & 0 & True & \textbf{72.6} & $\backslash$ & \textbf{41.3} & $\backslash$ & \textbf{91.0} & $\backslash$ \\
\midrule
19 & True & 2k & [0, 2, 4, ...] & 100 & True & \textbf{77.8} & \textbf{68.0} & \textbf{48.0} & \textbf{37.9} & \textbf{91.4} & \textbf{99.6} \\
\bottomrule
\end{tabular}
\end{table*}

\begin{table*}[ht]
\centering
\caption{The accuracy of Llama3.1-8B-Instruct on LongMemEval, LongBench-V2, and Ruler}
\label{tab:merged_llama}

\setlength{\tabcolsep}{2pt} 
\begin{tabular}{l|cclcc|c|c|c}
\toprule
\multirow{2}{*}{\textbf{No.}} & \multirow{2}{*}{\textbf{SFT}} & \textbf{Window} & \multirow{2}{*}{\textbf{FA layers}} & \textbf{Keep} & \textbf{FA} & \textbf{LongMem} & \textbf{LB-V2} & \textbf{Ruler} \\
 & & \textbf{size} & & \textbf{first} & \textbf{decode} & \textbf{Non-Think} & \textbf{Non-Think} & \textbf{Non-Think} \\
\midrule
0 & False & Full & / & / & / & \textbf{61.0} & \textbf{33.2} & \textbf{82.4} \\
1 & False & 2k & [] & 0 & False & 0.6 & 0.0 & 0.0 \\
2 & False & 8k & [] & 0 & False & 1.2 & 0.0 & 0.0 \\
\midrule
3 & False & 2k & [] & 10 & False & 1.8 & 28.9 & 0.0 \\
4 & False & 2k & [] & 0 & True & 0.0 & 0.0 & 0.0 \\
5 & False & 2k & [0, 2, 4, ...] & 0 & False & 3.0 & 0.0 & 0.0 \\
\midrule
6 & False & 2k & [] & 10 & True & 16.8 & 28.9 & 1.0 \\
7 & False & 2k & [] & 100 & True & 20.0 & \textbf{30.2} & 10.0 \\
8 & False & 2k & [] & 1k & True & 24.2 & \textbf{30.2} & 18.4 \\
9 & False & 4k & [] & 10 & True & 23.8 & 27.5 & 2.4 \\
\midrule
10 & False & 2k & [0, 2, 4, ...] & 10 & True & \textbf{42.6} & 28.2 & 22.8 \\
11 & False & 2k & [0, 4, 8, ...] & 10 & True & 17.8 & \textbf{32.6} & 4.4 \\
12 & False & 2k & [1, 3, 5, ...] & 10 & True & 21.0 & 26.5 & 23.4 \\
13 & False & 2k & [2, 6, 10, ...] & 10 & True & 24.4 & 26.8 & 1.0 \\
14 & False & 4k & [0, 2, 4, ...] & 100 & True & \textbf{44.0} & \textbf{30.9} & \textbf{52.4} \\
\bottomrule
\end{tabular}
\end{table*}




\begin{table*}[ht]
    \centering
    \caption{Efficiency metrics of different SWAA configurations on vLLM. "FA layers = 1/4" means one fourth of total layers use full attention while the others use SWA.}
    \label{tab:speed}
\setlength{\tabcolsep}{2pt} 

\begin{tabular}{cccc|ccc}
\toprule
\textbf{window} & \textbf{keep first} & \textbf{FA decode} & \textbf{FA layers} & \textbf{TTFT (s)} & \textbf{TPOT (s)} & \textbf{Throughput (k tks/s)} \\
\midrule
Full & 0 & False & None & 1681.44 & 0.16 & 3.74 \\
\midrule
2k & 0 & False & None & 203.20 & 0.02 & 30.72 \\
2k & 100 & False & None & 207.74 & 0.02 & 30.65 \\
2k & 0 & False & 1/2 & 938.00 & 0.09 & 6.70 \\
2k & 0 & True & None & 963.39 & 0.11 & 6.39 \\
2k & 0 & True & 1/2 & 1321.39 & 0.14 & 4.72 \\
2k & 0 & True & 1/4 & 1141.66 & 0.12 & 5.43 \\
\midrule
4k & 0 & False & None & 233.07 & 0.02 & 27.03 \\
4k & 100 & False & None & 237.87 & 0.02 & 26.74 \\
4k & 0 & False & 1/2 & 949.02 & 0.09 & 6.64 \\
4k & 0 & True & None & 990.00 & 0.11 & 6.23 \\
4k & 0 & True & 1/2 & 1340.91 & 0.14 & 4.64 \\
4k & 0 & True & 1/4 & 1166.69 & 0.13 & 5.32 \\
\bottomrule
\end{tabular}
\end{table*}

\begin{table*}[ht]
\centering
\caption{Results of different training epochs of Qwen3-4B-Thinking on LongMemEval}
\label{tab:epoch}

\begin{tabular}{c|cccc|c}
\toprule
 \textbf{SFT (epochs)} & \textbf{window size} & \textbf{FA layers} & \textbf{keep first} & \textbf{FA decode} & \textbf{Acc} \\
\midrule
 1 & 2k & [] & 0 & True & 58.0 \\
 2 & 2k & [] & 0 & True & 57.6 \\
 3 & 2k & [] & 0 & True & 56.0 \\
\bottomrule
\end{tabular}

\end{table*}
\begin{table*}[ht]
\centering
\caption{Results of LightTransfer on LongMemEval. "lazy" represents the half layers with higher lazy ratio, i.e. those which should apply SWA in theory. "non-lazy" represents the other part, i.e. those which should keep full attention.}
\label{tab:lazy}

\begin{tabular}{cclcc|cc}
\toprule
 \textbf{SFT} & \textbf{window size} & \textbf{FA layers} & \textbf{keep first} & \textbf{FA decode} & \textbf{Acc think} & \textbf{Acc non-think} \\
\midrule
\multicolumn{7}{l}{\textbf{Model Group: Qwen3-4B}} \\
\midrule
 False & 2k & [0, 2, 4, ...] & 100 & True & 48.8 & 18.4 \\
  False & 2k & [1, 3, 5, ...] & 100 & True & \textbf{70.8} & \textbf{50.4} \\
  False & 2k & lazy & 100 & True & 70.2 & 47.8 \\
  False & 2k & non-lazy & 100 & True & 54.0 & 19.6 \\
\midrule
\multicolumn{7}{l}{\textbf{Model Group: Qwen3-30B-A3B}} \\
\midrule
  False & 2k & [0, 2, 4, ...] & 100 & True & \textbf{75.8} & \textbf{64.2} \\
  False & 2k & [1, 3, 5, ...] & 100 & True & 60.2 & 25.8 \\
  False & 2k & lazy & 100 & True & 61.8 & 25.2 \\
  False & 2k & non-lazy & 100 & True & 74.8 & 59.2 \\

\midrule
\multicolumn{7}{l}{\textbf{Model Group: Llama3.1-8B-Instruct}} \\
\midrule
  False & 2k & [0, 2, 4, ...] & 100 & True & $\backslash$ & 39.8 \\
  False & 2k & [1, 3, 5, ...] & 100 & True & $\backslash$ & 24.2 \\
  False & 2k & lazy & 100 & True & $\backslash$ & 20.2 \\
False & 2k & non-lazy & 100 & True & $\backslash$ & \textbf{49.8} \\

\bottomrule
\end{tabular}

\end{table*}

\end{document}